\documentclass[conference]{IEEEtran}

\usepackage{amsmath}
\usepackage[dvips]{graphicx}
\usepackage{epsfig}
\usepackage{tikz, graphicx}
\usepackage{amssymb}
\usepackage{multirow}
\usepackage{float}
\usepackage{bm}
\usepackage{color}
\usepackage[lofdepth,lotdepth,caption=false,font=small]{subfig}
\usepackage{booktabs}
\usepackage{array}
\usepackage{lettrine}
\usepackage{type1cm}
\usepackage{gensymb}

\usepackage{fancyhdr}
\usepackage{hyperref}

\usepackage{algorithm}
\usepackage{algpseudocode}
\usepackage{varwidth}

\usepackage{tabularx}

\usepackage{xcolor}

\def\h#1{\skip255=#1truemm \divide\skip255 by 10
          \hskip\skip255}
\def\hh#1{\skip255=#1truemm \divide\skip255 by 100
          \kern\skip255}
\def\v#1{\skip255=#1truemm \divide\skip255 by 10
          \vskip\skip255}

\author{\IEEEauthorblockN{Federico Pittino\IEEEauthorrefmark{1}, Roberto Diversi\IEEEauthorrefmark{1}, Luca Benini\IEEEauthorrefmark{1}\IEEEauthorrefmark{2}, Andrea Bartolini\IEEEauthorrefmark{1}}
\IEEEauthorblockA{\IEEEauthorrefmark{1}Department of Electrical, Electronic and Information Engineering (DEI), University of Bologna, Italy \\\{federico.pittino, roberto.diversi, luca.benini, a.bartolini\}@unibo.it}
\IEEEauthorblockA{\IEEEauthorrefmark{2}Integrated Systems Laboratory, ETH Zurich, Switzerland \{lbenini\}@iis.ee.ethz.ch}}

\begin{document}

\bstctlcite{IEEEexample:BSTcontrol}

\title{Robust identification of thermal models for in-production High-Performance-Computing clusters with machine learning-based data selection}

\maketitle
\vspace*{-0.5cm}

\begin{abstract}
Power and thermal management are critical components of High-Performance-Computing (HPC) systems, due to their high power density and large total power consumption.
The assessment of thermal dissipation by means of compact models directly from the thermal response of the final device enables more robust and precise thermal control strategies as well as automated diagnosis. 
However, when dealing with large scale systems ``in production", the accuracy of learned thermal models depends on the dynamics of the power excitation, which depends also on the executed workload, and measurement nonidealities, such as quantization.
In this paper we show that, using an advanced system identification algorithm, we are able to generate very accurate thermal models (average error lower than our sensors quantization step of 1$^{\circ}$C) for a large scale HPC system on real workloads for very long time periods. However, we also show that: 1) not all real workloads allow for the identification of a good model; 2) starting from the theory of system identification it is very difficult to evaluate if a trace
of data leads to a good estimated model. We then propose and validate a set of techniques based on machine learning and deep learning algorithms for the choice of data traces to be used for model identification. We also show that deep learning techniques are absolutely necessary to correctly choose such traces up to 96\% of the times.
\end{abstract}

\vspace*{-0.1cm}

\section{Introduction} \label{sec:intro}

High performance computing (HPC) systems are designed to be at the cutting edge of computing capabilities. 
To achieve this goal, HPC installations are characterized by high computational power density and as a consequence, by high power density as well as large total power consumption. Indeed HPC systems have 2-4x higher rack power density w.r.t. server and industrial datacentre installations, with a per rack power envelope ranging between 20-100 kWatts \cite{Gilly_2016}. 
High power density and envelope are obviously critical for HPC system management and operation.

Today one of the most powerful supercomputers in Top500 is Sunway TaihuLight which consumes 15.3 MW for delivering 93 Petaflops. One of the previous first ones, Tianhe-2, consumes 17.8 MW for ``only'' 33.2 Petaflops. However, the power consumption increases to 24 MW when considering also the cooling infrastructure\cite{dongarra2013visit}. Such an amount of cooling power serves to prevent thermal issues. In fact, the performance of the processing elements is actively controlled by the internal firmware logic, which modulates chip voltage and frequency for maximizing the clock speed while satisfying power and thermal constraints. However these mechanisms are usually reactive, threshold-based and take significant safety margins: authors in \cite{Moskovsky:2016:SLL:3026773.3026777} show that, for hot-water liquid cooled nodes, the processors are incapable of employing thermal throttling by using DVFS states to prevent the critical thermal threshold to be reached. 

To solve these issues, several works in the literature \cite{Verma:2008:PDP:1375527.1375555,Rodero2012,zanini2013online,beneventi2016cooling,JMPC12,TCAS_Mutapcic_2009,reda2018blind} propose to take advantage of proactive thermal and power management strategies. These strategies all rely on the availability of compact predictive power and thermal models, capable of predicting future power consumption and temperature of the system and, even more importantly, to build a clear understanding on the sensitivity of these on workload parameters and hardware knobs that can be controlled at run time. Such models allow to estimate and model the power consumption of the entire CPUs and their cores based on workload characteristics extracted through performance counters and micro-architectural usage. Thanks to that, an optimizer can leverage these models to find the maximum clock frequency to apply based on the current usage of the micro-architecture while satisfying a global power budget or thermal constraints. 


Compact models can be used in combination with optimization and artificial intelligence techniques to select in a robust fashion the optimal operating points from the target power and temperature and the current conditions \cite{Verma:2008:PDP:1375527.1375555,Rodero2012,zanini2013online,beneventi2016cooling,JMPC12}. Moreover, such compact models can be used also for detecting anomalous changes in the behaviour of the system, for example due to a failure. However, the strategies for learning these models rely on design-time parameters that cannot cope with manufacturing variability, which makes each chip different from the others. Moreover, differences in deployment conditions and aging which can induce very significant differences in compact model parameters even for nominally identical nodes. In addition, such models have been applied only to single node systems operating in a test environment and therefore cannot easily be utilized at the scale of a full system in production without causing significant calibration costs of all its nodes (e.g. bringing the HPC machine off-line periodically for power and thermal characterization). 

In case an identified compact thermal model needs to be periodically updated during the lifetime of the system, additional challenges arise concerned with identifying models under production workloads, which cannot be calibrated to the needs of the model identification process. In fact, it is well known that the input signals may affect significantly the results of an identification experiment \cite{SodSto,Ljung}. The choice of a suitable input signal (when possible) has been extensively treated in the system identification literature \cite{SodSto,Ljung}. In many identification methods the input data are not available but the type of input excitation is assumed to be known \cite{TCAS_Blind_2005,TCAS_Blind_2007,TCAS_ARMA_2008}. When dealing with real workloads, as in our case, it is critical to evaluate if a given input (workload) will lead to a sufficiently accurate estimated model. However, even if \cite{SodSto,Ljung} suggest that persistently exciting of sufficiently high order as well as a low condition number is a requirement for high estimation accuracy, none of previous works have verified if these conditions are sufficient for discriminating real workload traces that lead to good or bad identified models.

This paper aims at bridging the gap between the theoretical identification of thermal models and their application to real workloads on in-production large HPC clusters. The paper is organized as follows. Sec. \ref{sec:related_work} presents a review with the state-of-the-art and highlights the open problems. Sec. \ref{sec:methods} describes the theoretical foundations of our system identification models. Sec. \ref{sec:results} describes our deployment scenario, consisting of a large HPC cluster under real workloads, and presents the results of thermal identification on our data. Given the large amounts of available data (14 days of continuous operation), we divide it in time windows of 12 hours each, which also enables us to highlight the problem of selecting the right window of data for good model identification. Finally, Sec. \ref{sec:winsel} explores the problem of window selection and it shows that only by means of machine learning and deep learning algorithms an accurate selection can be performed.

\section{Related work} \label{sec:related_work}

Several strategies have been proposed in the last decade for extracting compact thermal models directly from a processor chip's thermal response to a given power/workload stress input \cite{YWang,Cochran10,Coskun09,DaCheng12,JMPC12,JTM,TCAS2014,DATE2013,IECON2016,reda2018blind}. 
The simplest ones do not account for the multimodal nature of the thermal transient caused by the different building materials and their relative time-constants (i.e. die, heat-spreader and heatsink) \cite{JMPC12}. In \cite{YWang,Cochran10}, a first order dynamic thermal model is identified by solving a linear Least Squares (LS) optimization problem. Sharifi et al.~\cite{TajanaSS} show that, when a model is available, it can be used effectively to filter out measurement noise using a Kalman filter. 

Coskun et al.~\cite{Coskun09} use an Auto-Regressive Moving Average (ARMA) technique for predicting the future thermal evolution of each core. The derived model predicts future temperature by using only its previous values. Since it does not account directly for workload-to-power dependency, a Sequential Probability Ratio Test (SPRT) technique is used to rapidly detect changes in the statistical residual distribution (average, variance) and, then, to re-train the model, when it is no longer accurate.
Juan et al.~\cite{DaCheng12} uses a combination of a K-means clustering and an Auto-Regressive (AR) model to learn a compact model for fast thermal simulation. This approach is effective only when starting from a highly accurate thermal model of the HW. Moreover, the missing exogenous terms in both of the above approaches leads to neglecting the direct link between dissipated power and temperature. Then, the learning of a large number of different models is required to capture the characteristics of different functional units and program phases.
Reda et al.~\cite{reda2018blind} propose a method for estimating thermal models and power consumption only from the measurements of thermal sensors and total power consumption. Without making a-priori assumption on the core's power consumption as well as model structure/parameters. The results are promising but, as in other works, to correctly identify the thermal model it is required to excite the system with a controlled input (power steps long enough to reach steady state conditions).
Bartolini et al.~\cite{JMPC12} present a distributed model learning approach based on a set of Auto-Regressive eXogenous (ARX) models. Each core executes its own model learning routine generating a local thermal model. The model is used internally, in each core, by a local model-predictive controller. However this approach has been applied only to simulated systems and it is based on the assumption that per-core power traces and thermal sensor outputs are accurate and without noise.

Indeed, standard ARX models are suitable to represent the so-called ``process noise''\footnote{this is a stochastic process usually injected as additional (unknown) input in order to represent unavoidable model approximations}, but are based on the assumption that input and output data are accurate and not affected by measurement noise~\cite{Ljung,SodSto}.
Beneventi et al.~\cite{JTM} present an Output Error system identification strategy that is robust to quantization noise on the input temperature measurements. This is achieved by adding to the basic optimization problem a set of linear constraints that filter out the model parameters that are not physically valid. The approach is validated on a quad-core server platform. Unfortunately, the proposed methodology cannot handle ``process noise''. 

Previous works have shown that the relation between the core temperature and the dissipated power can be described by a purely dynamic ARX model \cite{DATE2013,TCAS2014}.
ARX models are widely used in system identification since they constitute the simplest way of representing a dynamic process in the presence of uncertainties \cite{Ljung}. Two important features of these models are the possibility of obtaining asymptotically unbiased estimates of their parameters by means of least squares and the absence of stability problems of the associated optimal one step-ahead predictors \cite{Ljung}. Nevertheless, it has been shown in \cite{TCAS2014} that the classic MISO ARX model is not able to describe properly the thermal dynamics of the system because the estimated models are characterized by relevant negative poles and/or complex conjugate poles. This is in contrast with the physics of thermal systems, where only real positive poles can exist. As explained in \cite{TCAS2014}, this problem is due to the presence of a significant level of measurement noise. 

To take into account the presence of this noise, MISO ARX models with noisy input and output have been considered \cite{DATE2013,TCAS2014}. These models belong to the family of errors-in-variables models and cannot be identified by means of standard least squares and prediction error methods \cite{EJC2010}.
In \cite{DATE2013} Diversi et al. introduced a bias compensated least squares approach for identifying noisy ARX models, which has been extended in \cite{TCAS2014} with a distributed implementation. These works have been conducted on the Intel Single Chip Cloud computer test device which featured ``cheap" ring oscillators as thermal sensors. In \cite{IECON2016} a Frisch scheme-based approach is applied to a server class processor operating in free-cooling with variable ambient temperature and the built-in state-of-the-art thermal sensors are affected by quantization noise. The obtained results prove the robustness of the approach.

To extract the models all the above mentioned works \cite{DATE2013,TCAS2014,IECON2016} rely on the capability of testing the system with Pseudo Random Binary Sequences (PRBS) workloads, where each core, synchronously with the thermal response measurements (with a regular sub-second sampling time) can be forced to execute at in ether a low workload/power (idle) state or high workload/power (power virus) state to emulate a Gaussian distribution of the power stimulus. The binary workload is chosen as it allows to pre-characterize precisely the power consumption of each of the two workload states and thus to create input vectors not affected by measurement noise in conjunction with an exciting workload.

In a realistic scenario, it is interesting to understand if the previously mentioned approaches can be applied to generic user workloads without a priori guarantees on the persistent excitation of the workload as well as whiteness of its spectral components. Indeed the system identification theory requires to excite the system under test in all its modes to be capable of correctly identify its model. However to track physical parameter changes as well as to detect abnormal changes in this parameters it is important to understand: (i) if real workloads, with no guarantees on their excitation, can be used to learn accurate models, and (ii) how to filter out workload windows which would not lead to accurate model identification. \\
Tackling these open problems is the main contribution of our work. In particular, point (i) above is discussed in Sec. \ref{sec:results}, while point (ii) is the topic of Sec. \ref{sec:winsel}.

\section{Methods} \label{sec:methods}

The objective of our work is the development and verification of a thermal model for an HPC cluster and its integration in a monitoring framework. Such a model is targeting the applications of thermal control and anomaly detection. As far as the requirements are concerned, it has to:
\begin{itemize}
    \item be predictive in time;
    \item have good accuracy and be stable in all operating conditions;
    \item provide temperature estimation at a core level, for most powerful control;
    \item be able to be estimated on real workloads, in order not to require down-time periods of the cluster for runs of ad-hoc workloads for model estimation.
\end{itemize}
In order to meet all these requirements, we have decided to extend the distributed and scalable monitoring framework that we presented in \cite{Beneventi2017, PittinoHPCS2018}. As far as the thermal model is concerned, we have used a variation to the algorithm presented in \cite{IECON2016}, which makes use of an identification algorithm (see Sec. \ref{sec:therm_mod}). \\
As we discussed, in our conditions several challenges arise:
\begin{itemize}
    \item the identification algorithms run under the assumption that the system to monitor is stressed with a very exciting (ideally white) input pattern; this condition is very likely to be not verified in real workloads;
    \item models identification should rely on the fact that there can be time windows where the above conditions for good model identification are verified, but it is not clear how to identify such windows.
\end{itemize}
In the following we describe in detail the derivation of the models and their integration into the monitoring framework \cite{Beneventi2017, PittinoHPCS2018}.

\subsection{Power model}

The thermal model \cite{IECON2016} uses as input the power of all cores in a package. Since in the architecture we considered there is no available measurement of the power per core, but only at a package level, the first step is to derive an estimate for the cores' powers. For this purpose, we have modified the model in \cite{PittinoHPCS2018} to obtain a power model at the core level and not at the package level. 

The power model is based on the measurement of appropriate metrics derived from the performance counters, as explained in \cite{PittinoHPCS2018}. To choose a subset of the metrics, among all the available ones, we have calculated Pearson's correlation of the metrics with the measured package power. We have then chosen the metrics with the highest correlation, and these are reported in Tab. \ref{tab:counters}. Note that the set comprises both core metrics and CPU metrics, because our power model needs to estimate also the power absorbed by the uncore. We have verified that these metrics are the smallest subset which gives good accuracy (a decrease from 97\% to 91\% for the number of points below the 9.7W threshold, see the results in Sec. \ref{sec:pow_model}), and that the addition of other metrics does not increase the model's performance considerably. All used metrics have also been rescaled in order to lie in the interval $[0,1]$, in order to avoid numerical instabilities in the model.

\begin{table}[tb]
\caption{Chosen metrics}
\label{tab:counters}
\begin{tabularx}{\columnwidth}{lX}
\hline
Metric name        & Description \\ \hline
freq $\cdot$ C0 & (MHz) Actual core frequency multiplied by ratio of time in state C0 \\
freq\_pkg & (MHz) Actual package frequency multiplied by ratio of time in state C0 \\
1 - C0\_pkg & Ratio of package time in all states except C0 \\ \hline
\end{tabularx}
\end{table}

The power model is then derived using a linear regression algorithm directly on the chosen metrics. It can be cast in the form:
\begin{align}
P_{pkg} = \sum_{i \in M_{unc}} \alpha_i m_i + \sum_{k \in C} \sum_{i \in M_{core}} \beta_{i,k} m_{i,k} \label{eq:lr_power}
\end{align}
where $M_{unc}$ and $M_{core}$ are the sets of uncore and core metrics, respectively, $C$ the set of cores, $m_i$ and $m_{i,k}$ the values of the metrics and $\alpha_i$ and $\beta_{i,k}$ the regression coefficients. Such a formulation allows us to partition the package predicted power in two sets of contributions:
\begin{align}
P_{pkg} = P_{unc} + \sum_{k \in C} P_k
\end{align}
with obvious definitions of the symbols comparing with Eq. \ref{eq:lr_power}. With this formulation we can easily extract the partial powers, i.e., $P_{unc}$ is an estimate of the power dissipated by the uncore and $P_k$ is the power dissipated by core $k$.

Note that also for the derivation of the power model we use real workloads, and this poses the same challenges we discussed above and in \cite{PittinoHPCS2018}. However, since the accuracy of the power model is not critical and the model itself is relatively simple, we have verified that we are always able to derive a model with relatively good stability and accuracy if we use a long enough interval for the training data, in our case generally 3 days as in \cite{PittinoHPCS2018}. Moreover, as we showed in \cite{PittinoHPCS2018}, the training of the power model and its evaluation can be done very efficiently in our scalable framework and they impose only a minimal overhead.

\subsection{Thermal model of a core}
\label{sec:therm_mod}

The thermal dynamics of a single core of a node are represented by means of a MISO model linking the core's temperature (model output) to the powers of all the node's cores (model inputs). As discussed in \cite{IECON2016}, a standard ARX (AutoRegressive eXogenous) model is not suitable to describe the thermal dynamics of a core because of the presence of a significant additive noise corrupting the temperature readings. Therefore, we will adopt the following MISO ARX model with additive output noise
\begin{gather}
\label{eqMISOARX}
	\bar T(t) + \sum_{i=1}^n a_i\,\bar T(t-i) = \sum_{k=0}^{N_c} \sum_{i=1}^n b_{ki}\,P_k(t-i) + w(t) \\
\label{eqMISOARX2}
	 T(t) = \bar T(t) + v(t),
\end{gather}
where
\begin{itemize}
\item[--] $\bar T(t)$ is the actual (unknown) core temperature;
\item[--] $n$ is the model order, i.e. the memory of the difference equation;
\item[--] $N_c$ is the number of cores of the node;
\item[--] $P_1, \dots, P_{N_c}$ are the dissipated powers of all the cores of the node and $P_0$ denotes the uncore power $P_{unc}$;
\item[--] $w(t)$ is the equation error (process noise), assumed to be a zero mean white process with variance $\sigma_w^2$;
\item[--] $T(t)$ is the measured core temperature;
\item[--] $v(t)$ is the additive measurement error, assumed to be a zero mean white process with variance $\sigma_v^2$, uncorrelated with $w(t)$ .
\end{itemize}
The identification problem to be solved consists in estimating the model coefficients $a_i, i=1,\dots,n$, $b_{ki}, k=0,\dots,N_c, i=1,\dots,n$ on the basis of $N$ samples of the measured temperature $T(t)$ and of the powers $P_j(t), j=0,\dots,N_c$. The equation error variance $\sigma_w^2$ and the additive noise variance $\sigma_v^2$ are also estimated.
\par
The adopted identification algorithm is an evolution of that presented in \cite{IECON2016} and is essentially based on the following equations
\begin{gather}
\label{eqsyst1}
     (\Sigma-\tilde\Sigma)\,\bar\theta = 0 \\
\label{eqsyst2}
     \Sigma_q\,\bar\theta = 0
\end{gather}
where
\begin{gather}
\label{eqtheta}
                \bar\theta = \bigl[\,1\,a_1\,\cdots\,a_n\,\,b_{01}\,\cdots\,b_{0n}\,\cdots\,\,b_{N_c1}\,\cdots\,b_{N_cn}\,\bigr]^T \\
                \noalign{\v{10}}
\label{eqSigmatilde}
	 \tilde\Sigma = {\rm diag}\,\,[\sigma_v^2 + \sigma_w^2\,\,\underbrace{\sigma_v^2\,\,\cdots\,\,\sigma_v^2}_n\,\,
                               \underbrace{0_{\phantom v}^{\phantom 2}\,\,\cdots\,\,0_{\phantom v}\h{-15}}_{(N_c+1)n}\h{5}] \\
\label{eqSigma}
		\Sigma = E\,[\varphi(t)\,{\varphi}^T(t)] \\
		\Sigma_q = E\,[\varphi_q(t)\,\varphi^T(t)]
\end{gather}
and
\begin{align}
  &\varphi(t) \h{-5}=\h{-5} [\h{5} -T(t)\,\dots\,-T(t-n)\, P_0(t-1)\,\dots\,P_0(t-n) \nonumber\\
  & P_1(t-1)\,\dots\,P_1(t-n)\,\dots\,P_{N_c}(t-1)\,\dots\,P_{N_c}(t-n)\h{5}]^T\\
\label{eqvarphiq}                 
 &\varphi_q(t) = [\h{5} P_0(t-1)\,\dots\,P_0(t-q)\,\,P_1(t-1)\,\dots\,P_1(t-q) \nonumber\\
  &\h{100}\dots\,P_{N_c}(t-1)\,\dots\,P_{N_c}(t-q)\h{5}]^T.
\end{align}
$E[\cdot]$ denotes the expectation operator. The integer $q$ in \eqref{eqvarphiq} is a user-chosen parameter. It can be noted that \eqref{eqsyst1}-\eqref{eqsyst2} is a system of equations where the unknowns are the model coefficients and the noise variances whereas the matrices $\Sigma$ and $\Sigma_q$ can be directly estimated from the available data: 
\begin{align}
\label{Sigmasamp}
	\hat\Sigma &= \dfrac{1} {N-n}\sum_{t=n+1}^{t=N}\h{-15}\varphi(t)\,\varphi^T(t)\\
	\hat\Sigma_q &= \dfrac{1} {N-q}\sum_{t=n+q+1}^{t=N}\h{-20}\varphi_q(t)\,\varphi^T(t).
\end{align}

\par
To apply the above mentioned identification approach, the sample matrices $\hat\Sigma$ and $\hat\Sigma_q$ need to be non singular. This implies that the dissipated powers (input signals) $P_0, P_1, \dots, P_{N_c}$ have to be persistently exciting of sufficiently high order. Nevertheless, as pointed out in \cite{TCAS2014}, the non-singularity of the matrices is often not sufficient to get satisfactory results. For this reason, the application of the algorithm proposed in \cite{IECON2016} can be difficult in the framework considered in this paper as we use real workloads and not ad-hoc excitations. This means that it is highly likely that the input is not highly exciting. In \cite{TCAS2014} some metrics (like the matrix condition number) have been proposed to evaluate the quality of the identified model, and in this work we test such assumptions. In the framework under study it is therefore necessary to include a block which performs the window selection procedure. More precisely, given a set of input-output data, the aim of this block is to evaluate the goodness of the set w.r.t. the implementation of the identification algorithm. 
\par
In order to evaluate the performance of the model, as in \cite{IECON2016}, we rely on a Kalman filter for the prediction of the temperature, using the identified thermal model. The filter is based on the following state space representation of the model  \eqref{eqMISOARX}--\eqref{eqMISOARX2}:
\begin{align}
\label{eqSSmodel1}
	x(t+1) &= A\,x(t) + B\,u(t) + G\,w(t+1) \\
\label{eqSSmodel2}
    T(t) &= C\,x(t) + v(t) = \bar T(t) + v(t) 	  
\end{align}
where
\begin{equation*}
\begin{aligned}
  A &= \begin{bmatrix}  -a_1 & 1 & 0 & \cdots & 0\, \\
                      -a_2 & 0 & \ddots & \ddots & \\
                      \vdots & \vdots & & \ddots & \\
                      \vdots & \vdots & & & 1\, \\
                      -a_n & 0 & \cdots & \cdots & 0\, \end{bmatrix} &
  B &= \begin{bmatrix} b_{01} & \cdots & b_{N_c1} \\
                      b_{02} & \cdots & b_{N_c2} \\
                      \vdots & & \vdots \\
                      \vdots & & \vdots \\
                      b_{0n} & \cdots & b_{N_cn} \end{bmatrix} \\
  C &= \begin{bmatrix} 1 & 0 & \cdots & 0 \end{bmatrix} & G &= C^T,                    
\end{aligned}
\end{equation*}
and
\begin{equation}
    u(t) = [\,P_0(t)\, P_1(t)\, \dots\, P_{N_c}(t)\,]^T.
\end{equation}
The filter allows to compute at each time step the prediction $\hat{\bar T}(t|t-1)$ of the actual core temperature $\bar T(t)$. This prediction can then be compared with the measured temperature $T(t)$ and the prediction error (innovation)
\begin{equation}
\label{filt_innov}
 \varepsilon(t) = T(t)-\hat{\bar T}(t|t-1)
\end{equation}
can be exploited as a model performance index. 
\par
\begin{figure}[tb]
	\centering
		\includegraphics[width=1\columnwidth]{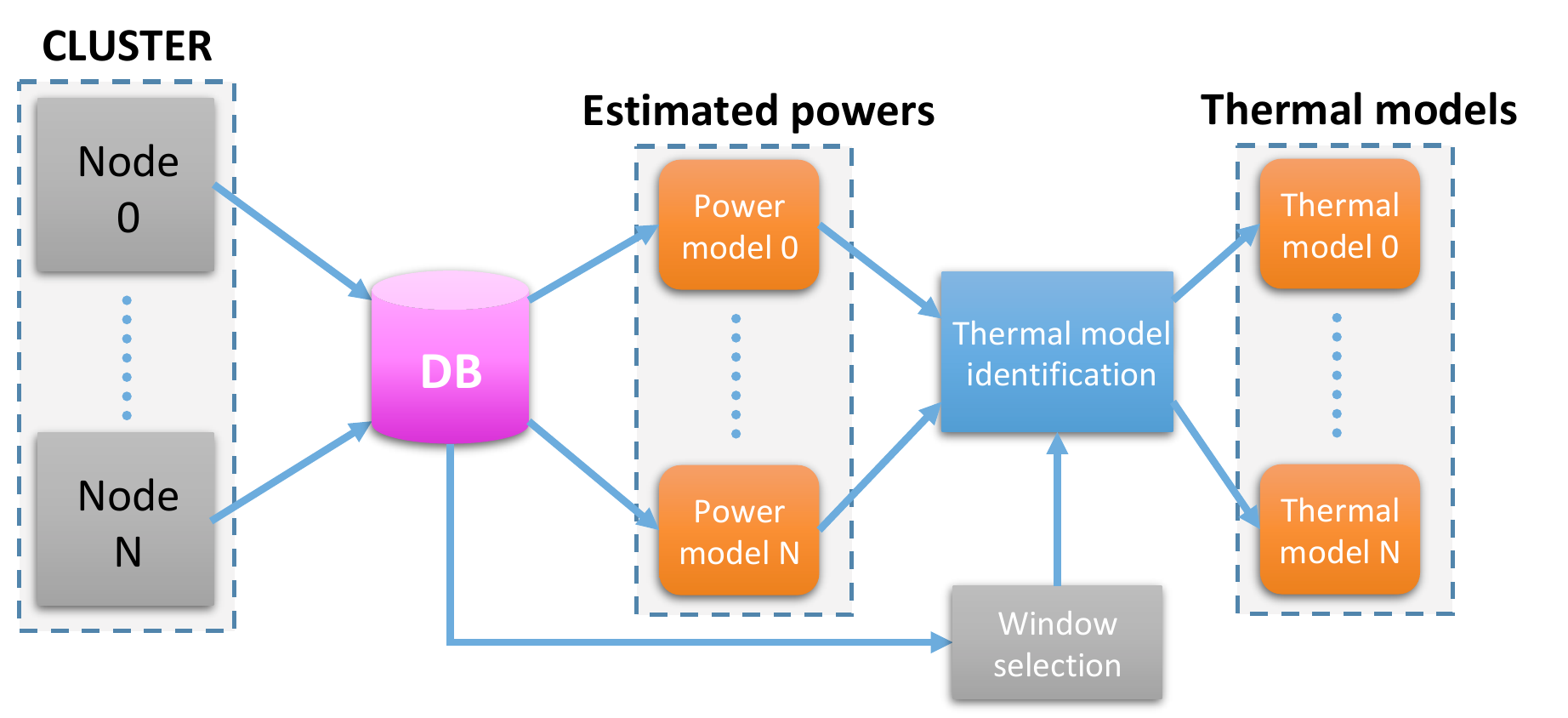}
		\vspace*{-0.5cm}
			\caption{Model-learning framework }
	\label{fig:modellearn}
\end{figure}

An overview of the framework architecture, where the power and thermal prediction modules and the window selection block are highlighted, is presented in Fig. \ref{fig:modellearn}.

\section{Results} \label{sec:results}

\subsection{Test bed} \label{sec:test_bed}

For our experimentation, we implemented the framework from Fig. \ref{fig:modellearn} on 40 of the 516 nodes of a cluster in a working production system (Galileo at CINECA \cite{GalileoCineca}) as a case study. Each node is equipped with two 8-cores Intel Haswell CPUs (E5-2630 v3 @ 2.40GHz) and 128GB of DRAM. The nodes have been monitored over a period of 17 days of normal operation, and all metrics are acquired at a constant sampling time of 2s. On the monitored nodes and during the entire time period there were 117 active users and a total of 3612 jobs were submitted, with an average of 31 users and 90 jobs per node (note that each user can submit multiple jobs, each using several nodes). The power prediction algorithms are instead run on a separate service node (Intel Haswell E5-2670 v3 @ 2.30GHz, 24 cores, 128GB DRAM, 4 NVIDIA GeForce GTX 1080 Ti), where the Apache Spark environment and all processing utilities are installed.

We want again to stress out the fact that, unlike most of the previous literature on power models (for example, \cite{BrooksPowModel2006}, \cite{IRITPowPredict2014}, \cite{PoznanPowModel2017}, \cite{SouthamptonPowModel2017}, \cite{TUDSouthamptonPowModel2017}) and our previous work on thermal models (\cite{IECON2016},  \cite{TCAS2014}), in this work we trained and applied our power and thermal prediction models in a production environment while the machine was fully operational and running user jobs, where each node has a different workload which can drastically change over time, and not on custom-defined workloads on single nodes.

\subsection{Power model} \label{sec:pow_model}

\begin{figure}[tb]
	\centering
		\includegraphics[width=1\columnwidth]{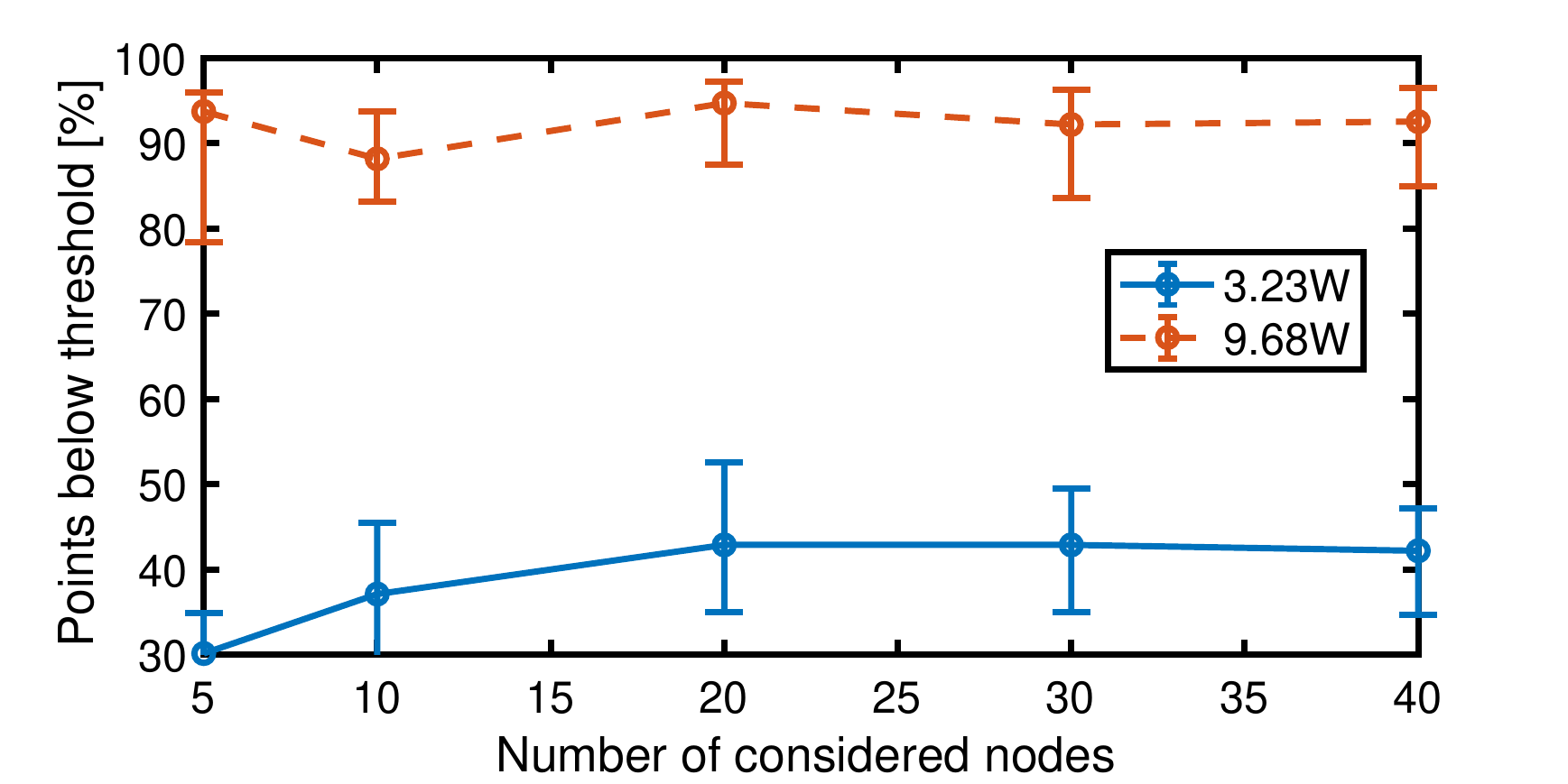}
			\caption{Overview of the error in estimating the total package power. The two lines show, between all nodes and sockets, the median (circle) and 25th and 75th percentiles (error bar) of the percentage of points where the power error due to the estimation is lower than a threshold, as we did in \cite{PittinoHPCS2018}.}
	\label{fig:PowModError}
\end{figure}

For the training of the power model, we have selected the first 3 days of operation of all nodes, and these data have not been used further in the definition of the thermal model. A different model is trained for each node and package, to reflect the variability between the nodes. \\
An overview of the model's performance is reported in Fig. \ref{fig:PowModError}, which shows the percentage of points where the error between the estimated and the measured power is below a certain threshold (3.23W and 9.68W, equivalent to the 1{\degree}C and 3{\degree}C errors we used in \cite{PittinoHPCS2018}). Each point on the x-axis is a different training time, and for each training time and threshold they are shown the median relative number of points below the threshold together with the 25\% and 75\% percentiles (the error bars). By comparing Fig. \ref{fig:PowModError} with the results in \cite{PittinoHPCS2018}, we note that the new model performs slightly worse than the old one, as expected because we use significantly less features and we are estimating also the individual cores contributions. However the additional error is not very significant, and anyways for more than 90\% of the points the error in package power is below 10W, which corresponds to roughly 1W error per core (assuming uniform distribution of the power) and about 10\% of the maximum power. As we will see in the following, these errors are indeed small and do not prevent us from obtaining very accurate thermal models.

\subsection{Thermal model}

For the identification and evaluation of the thermal models we have used the 14 days of measurements remaining from the power models training. To evaluate the performance of the proposed model on this dataset we have decided to divide this data in shorter time windows. One of the reasons behind this choice is to be comparable with previous works \cite{IECON2016} which have used relatively short windows ($\sim$1.5 hours). Another reason is that, since the workload can vary considerably during the entire time frame, having multiple windows available for each node enables us to perform a thorough cross-validation of our results. \\
On the other hand, the time windows cannot be too short, since the workloads have relatively large time constants and a short time window would not be able to easily capture all the possible states of the system. For all these reasons, we have chosen to divide the time frame in 25 windows, each $\sim$12 hours long.

\begin{figure}[tb]
	\centering
		\includegraphics[width=1\columnwidth]{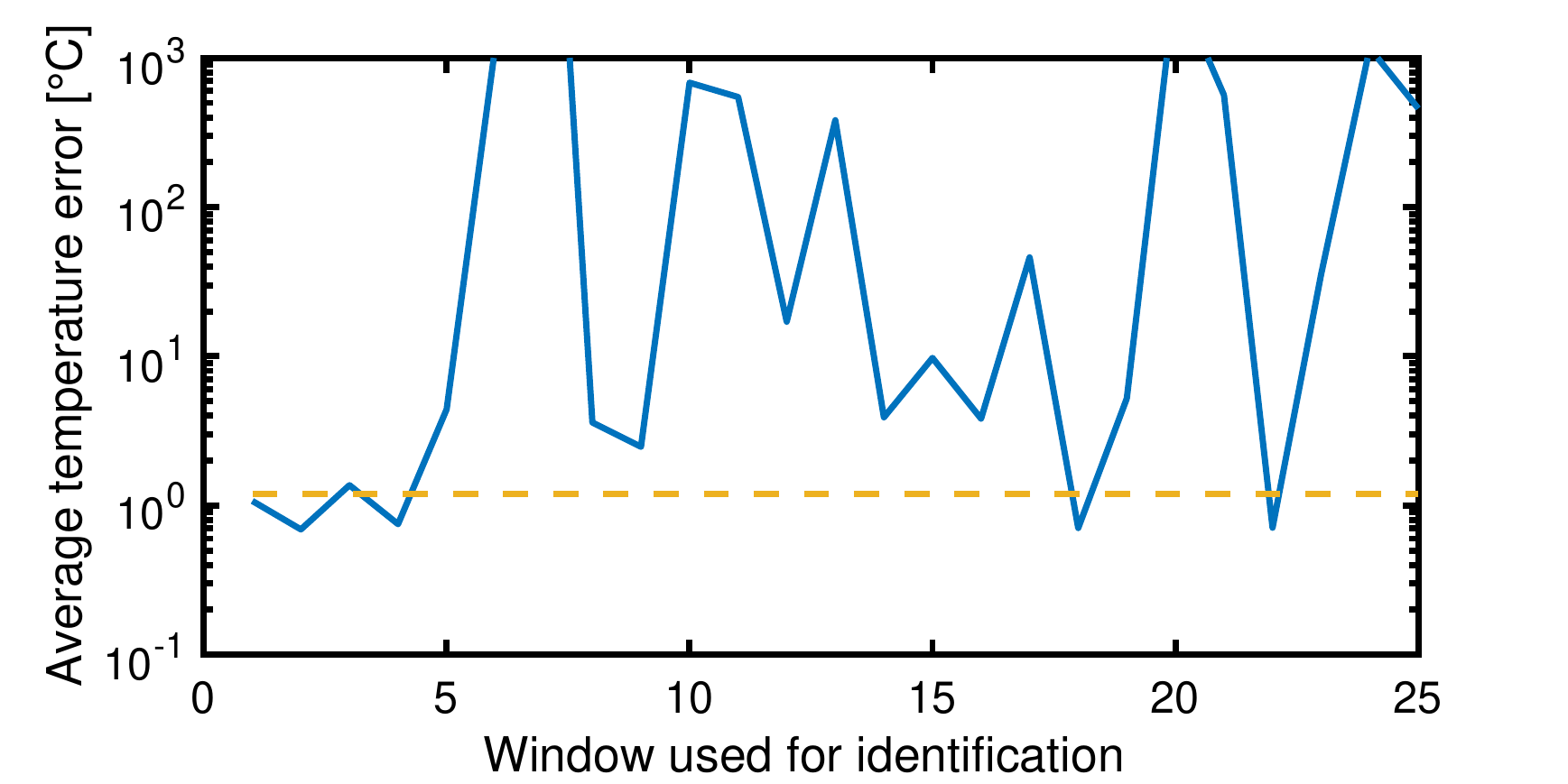}
			\caption{Average of the temperature error across all cores for one package in the cluster, changing the window used for the thermal model identification. The dashed horizontal line corresponds to 1.2$^\circ$C.}
	\label{fig:Terrmean}
\end{figure}

\begin{figure}[tb]
	\centering
		\subfloat[window 16]{\includegraphics[width=1\columnwidth]{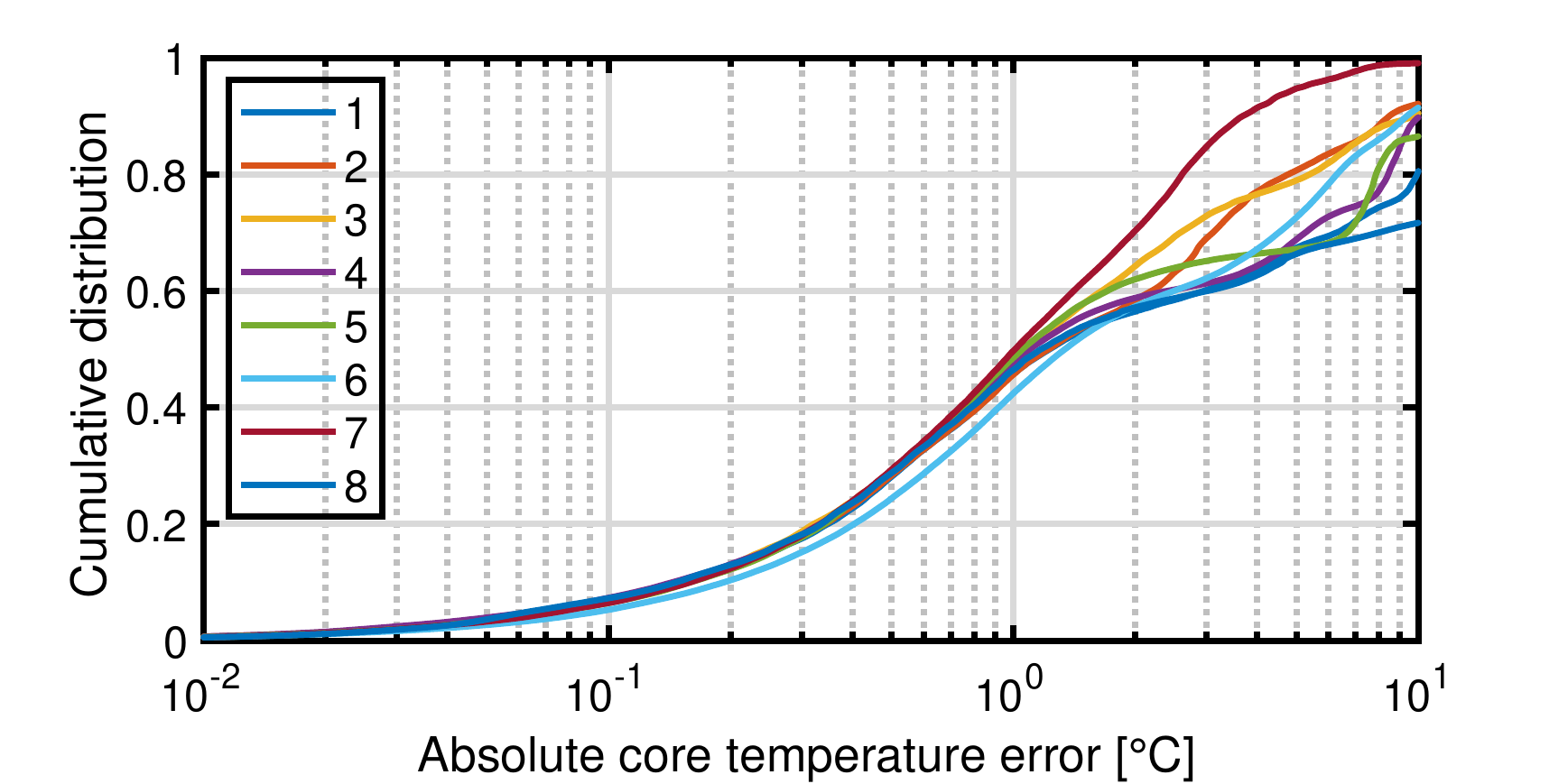}} \\
		\subfloat[window 22]{\includegraphics[width=1\columnwidth]{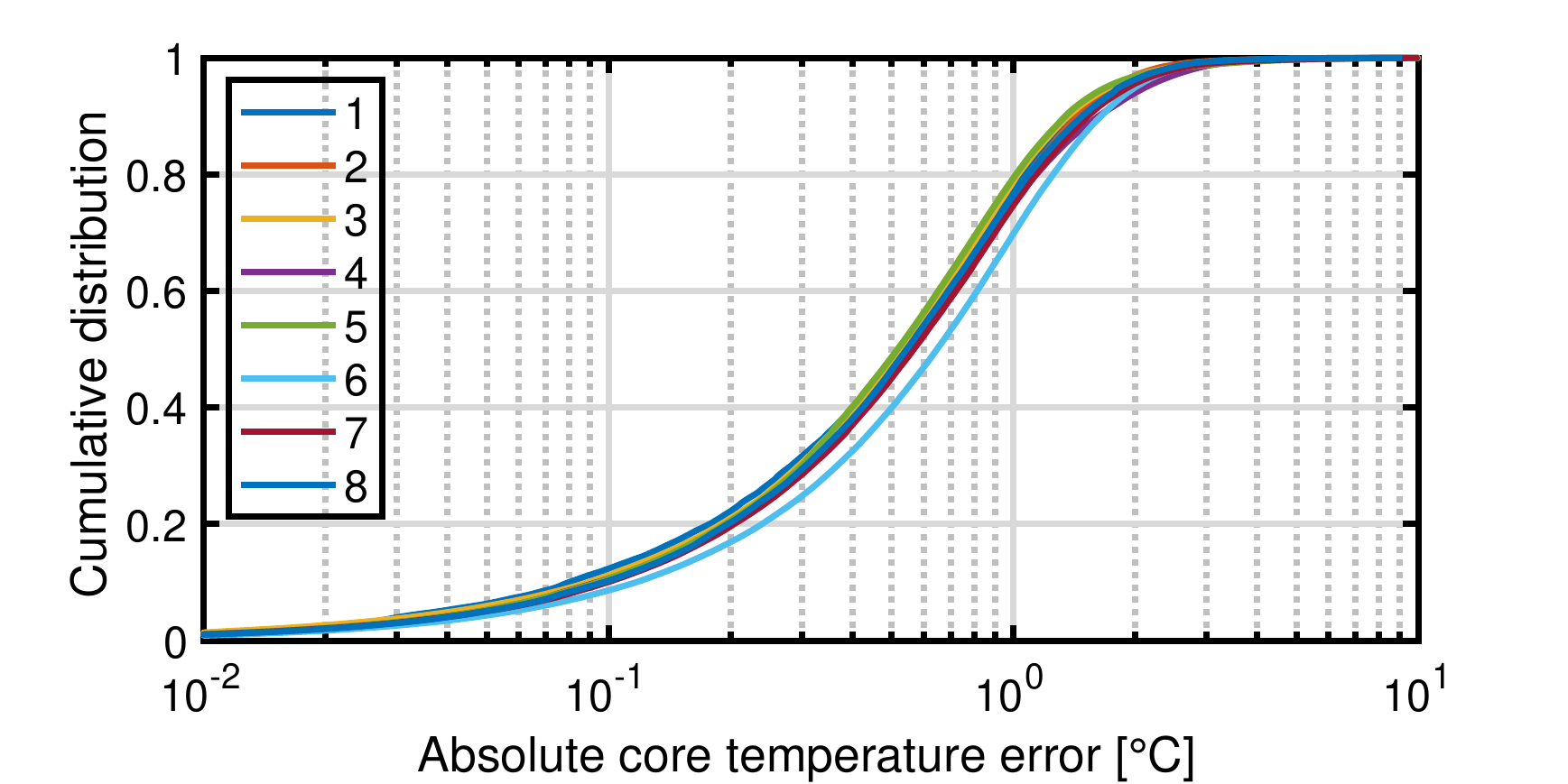}}
			\caption{Distribution of the temperature error for all cores of one package in the cluster, changing the window used for the thermal model identification.}
	\label{fig:Terrdistr}
\end{figure}

\begin{figure}[tb]
	\centering
		\subfloat[window 16]{\includegraphics[width=1\columnwidth]{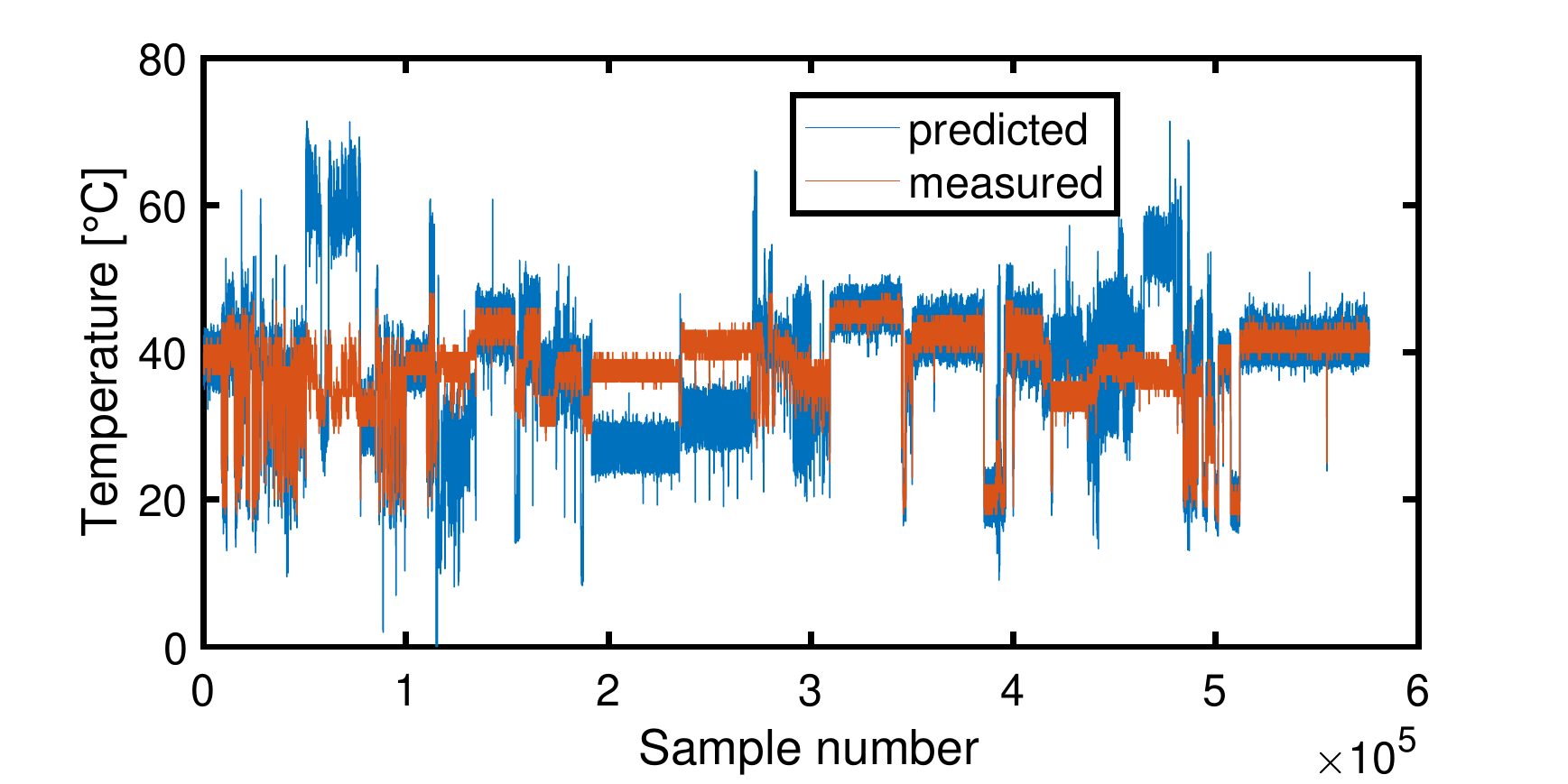}} \\
		\subfloat[window 22]{\includegraphics[width=1\columnwidth]{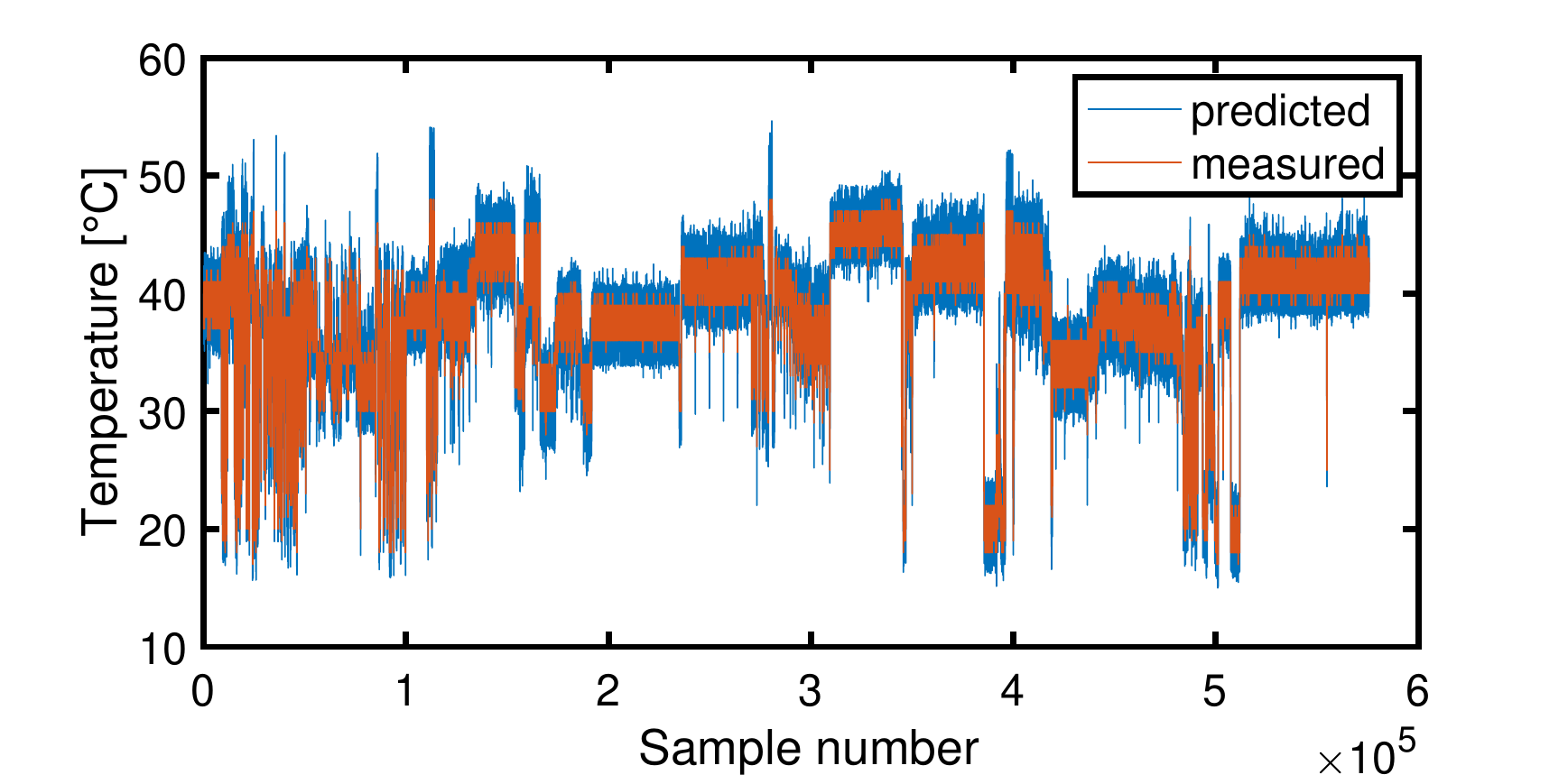}}
			\caption{Time traces of the measured and predicted temperatures for one core of one package in the cluster, changing the window used for the thermal model identification.}
	\label{fig:Tpred}
\end{figure}

For each core of a node, a MISO ARX model with additive output noise of order $n=2$ has been identified. The thermal dynamic of each core is thus characterized by two poles. The suitability of this choice has been proved in \cite{IECON2016}. The estimated model on each window is then been used to compute the temperature predictions on all the remaining data by means of a Kalman filter, as discussed in Sec. \ref{sec:therm_mod}. It is important to note that, to assess the model's performance, we have performed a very strict cross-validation, since we test the model on an amount of data which is 25 times larger than the one used for identification. Fig. \ref{fig:Terrmean} reports the average of the error between the predicted and measured temperatures across all cores in one package for the whole available time frame of 14 days, changing the window used for the thermal model identification. It is evident that there is a great variability between the models identified in different windows. In particular, some of the models have excellent performance, with an average error often below 1.2$^\circ$C (the dashed horizontal line), which is only slightly higher than the quantization step of 1$^\circ$C.

Going more in detail, Fig. \ref{fig:Terrdistr} shows the distribution of the error separately for all cores in the package choosing two windows for the identification, one with small error and one with high error. Moreover, Fig. \ref{fig:Tpred} chooses one of the cores and shows also the time trace of both the predicted and the measured error in these situations. It is once again evident that, if the model is identified correctly, the performance is excellent, while, if the window chosen for the identification has bad properties, applying a SoA system identification algorithm to real workload data can lead to poor estimations. It is therefore of utmost importance to devise a robust procedure for the a-priori evaluation of an identified model, or even for the choice of an appropriate window of data to use. This necessity becomes even more prominent when the models are used for control and anomaly detection.

\section{Window selection} \label{sec:winsel}

As we have discussed, an appropriate choice for the data window to use for identification is of paramount importance. Such a choice can be made in two ways:
\begin{enumerate}
\item a-priori, before the identification, basing the choice upon the properties of solely inputs (powers) and outputs (temperatures);
\item using the results of the identification, looking for example at the poles, the condition number or the whiteness of the residuals.
\end{enumerate}
Ideally, method 1) above should be preferable, as long as the algorithm for window selection is computationally very efficient. On the other hand, it is not guaranteed that a good window selection can be performed using only the results of the identification.

In order to evaluate the model's performance, we have derived a ground truth a-posteriori by using a Kalman filter on new data (see Sec. \ref{sec:therm_mod}). Such a method can not be used for the window selection, both because it is highly computationally intensive and also since it might lead to delays or problems in a system that is meant for control or anomaly detection. \\
We have decided to perform the window selection on a per-core basis, i.e., every core in the package has its own ensemble of chosen windows, that can be (and likely in most cases will be) shared across most cores in one package. Note that the thermal model for one core uses as input also the estimated powers of all other cores and of the uncore in the package, so we expect that a selection on a per-core basis will still take into account the relative difference in activity between the cores.

In the following we test multiple algorithms on both methods for window selection. All the algorithms lie in the field of supervised machine learning, and therefore need to define labeled training and test sets. The goal of the algorithms is to perform a classification, calculating a likelihood of the windows to be good (windows whose identified model has good performance) or bad ones. \\
The data used for training of all algorithms has been handled in the same way. All 25 windows per package have been divided by core, resulting in 16k total windows. These have been randomly and independently reordered, and the 80\% of those has been used for the training phase, the remaining for the test phase. A window has been labelled as good if the average error of the Kalman filter on the entire data using the parameters identified in that window is below a threshold (1.2$^\circ$C in our case), its standard deviation is also below a threshold (1.5$^\circ$C) and the poles are stable, real and not too low (we set a threshold of 0.8 based on Fig. \ref{fig:ThermmodIdentMetrics} and on the results in \cite{IECON2016}). For better robustness of the algorithm, we have excluded from the training set the windows with medium temperature error (between 1.2$^\circ$C and 1.5$^\circ$C) and standard deviation (between 1.5$^\circ$C and 2$^\circ$C). All these thresholds have been calibrated by inspection of the data, and the threshold on the error standard deviation is necessary to exclude models where the error distribution deviates too much from a normal distribution (see for example Fig. \ref{fig:Terrdistr}). 

The performance has been evaluated by classifying the windows from the test set, identifying the predicted good windows (i.e., the ones where the computed likelihood is above a given threshold) and showing the average temperature error that we would incur if we chose that window for the identification. Note that the misclassified good windows (i.e., the windows that are predicted of being good while in fact they are not) are the worst-case scenario, since in this case we would use for temperature control a bad model, while the misclassified bad windows are not as critical.

\subsection{Selection based on time traces}

For the selection based on time traces, we have then chosen to use as input the time traces of the cores' and the uncore partial powers (as estimated by the model from Sec. \ref{sec:pow_model}) and the core measured temperature. Since we have observed in the previous section that the identified model performs badly if the poles are very different between the cores, we have also decided to include the time trace of the measured package power, in order to obtain a measurement of similarity between all cores in a package. 

We have evaluated three choices of algorithms, which will be presented in detail in the following part of this section. In summary, two of them rely on custom-defined features calculated on the time traces (namely, a classical SVM with RBF kernel \cite{SVM1995} and a fully connected neural network \cite{DeepLearningBook2016}) and the last one is a 1D Convolutional Neural Network (CNN) \cite{DeepLearningBook2016} applied directly to the time traces. All neural network models have been implemented in PyTorch \cite{PyTorch2017}, levering the multi-GPU capability of our system.

As far as testing is concerned, the output of the neural networks is a number in the interval $[0,1]$ which represents the likelihood of the window being good. If we set the decision threshold at 0.5, all windows will be classified either good or bad. We can also decide to use a different threshold, for example 0.8, and in this case only windows with likelihood $l> 0.8$  are considered good. As for the remaining windows, we have decided to leave all the ones with likelihood in the range $[0.2, 0.8]$ are unclassified. This allows us to define three classes: (i) good windows $l> 0.8$, to be used for identification, (ii) bad windows $l < 0.2$, to be discarded, (iii) unclassified windows $0.2 < l < 0.8$, for which we are not sure whether they can be used or not, and they can be a pool to choose from if no window gets classified as good.

\subsubsection*{Classification based on custom features}

\begin{figure}[tb]
	\centering
		\subfloat[NN trace]{\includegraphics[width=1\columnwidth]{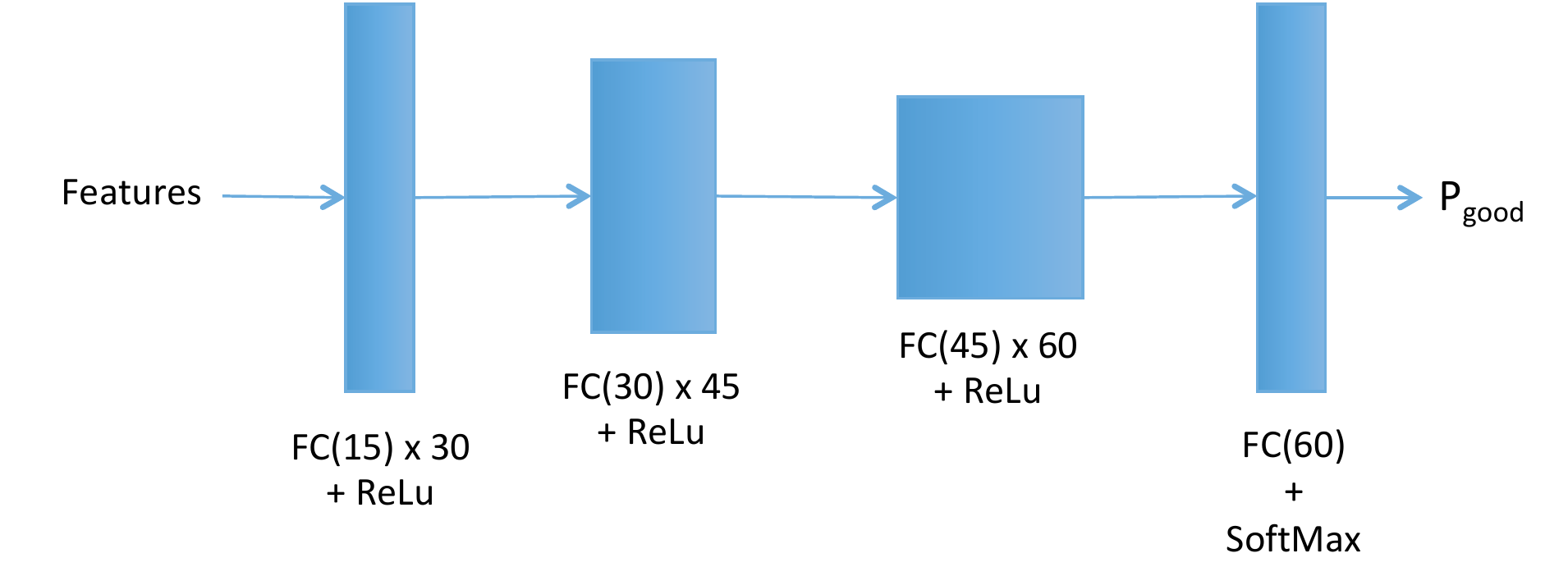} \label{fig:MLCoreNet_architecture}} \\
		\subfloat[CNN trace]{\includegraphics[width=1\columnwidth]{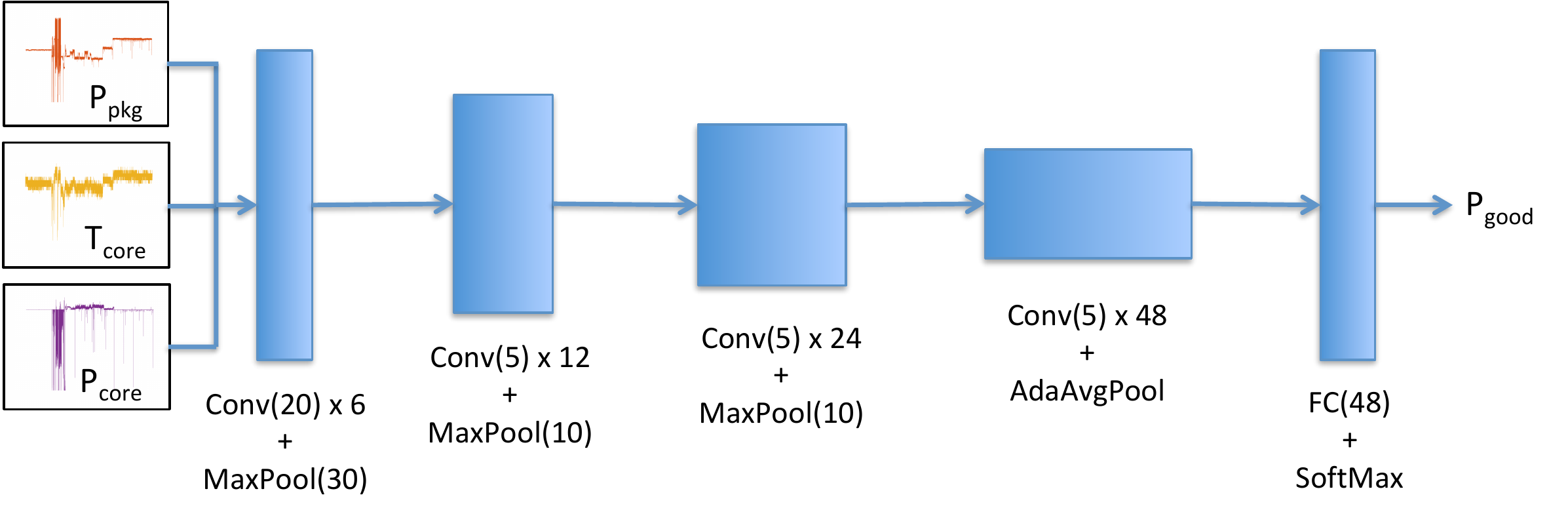} \label{fig:WinCoreNet_architecture}}  \\
		\subfloat[NN identification]{\includegraphics[width=1\columnwidth]{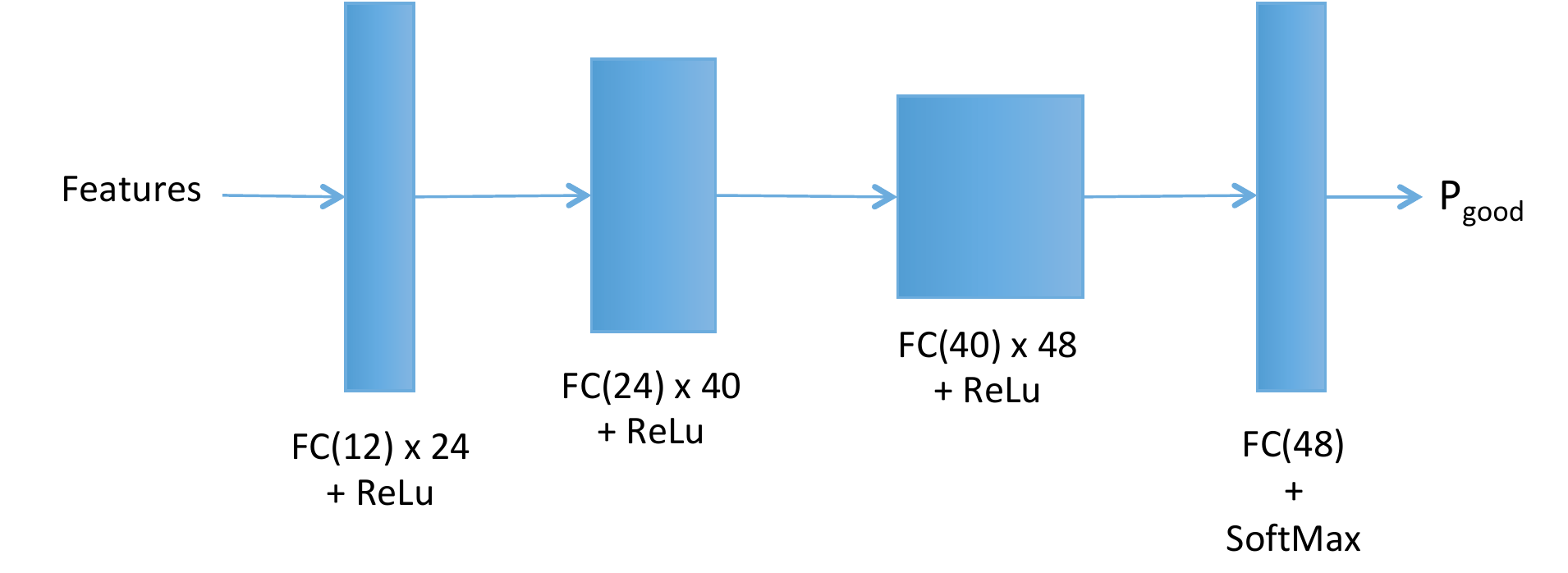} \label{fig:MLCoreIdentNet_architecture}} 
			\caption{Architecture of the Neural Networks used for window selection.}
\end{figure}

\begin{table}[tb]
\caption{Chosen features for the SVM and NN algorithms using the time traces, where the signal $x$ can stand either for package power or core's temperature or partial power.}
\label{tab:SVMfeatures}
\begin{tabularx}{\columnwidth}{lX}
\hline
Feature definition        & Description \\ \hline
$\text{E}\left[\left|\text{corr}\left(x\right)\right|\right]$ & Average of the absolute value of the signal's autocorrelation with up to 100 samples delay \\
$\text{E}\left[\bar{x}\right]$ & Average of the low-passed signal \\
$\text{std}\left[\bar{x}\right]$ & Standard deviation of the low-passed signal \\
$\max\left[\bar{x}\right] - \min\left[\bar{x}\right]$ & Maximum span of the low-passed signal \\
$\text{E}\left[\left\|\text{fft}\left(x\right)\right\|\right]$ & Average of the norm of the signal's FFT \\ \hline
\end{tabularx}
\end{table}

As already outlined, we have used both a SVM algorithm with radial basis functions and a fully connected neural network. The architecture of the network is shown in Fig. \ref{fig:MLCoreNet_architecture} and it has been derived following the standard suggestions for neural networks \cite{DeepLearningBook2016}. It consists of four fully connected layers each followed by a rectified nonlinearity (ReLu) and finally a softmax classification layer. The optimizer used for training is Adam, with learning rate 0.001 and weight decay $10^{-4}$ and using as loss function the binary cross-entropy. The training phase was stopped after 3000 iterations, where the training loss had converged to a value of 0.35. 

Both algorithms use the same input, which consists in a pre-defined set of features on the power and temperature time traces. The features have been chosen in order to empirically reproduce the characteristics that we have observed in good windows. Some of the features are defined on a low-pass version of the traces (denoted as $\bar{x}$ for a signal $x$), where the window signal has been further divided into 20 subwindows and averaged. Tab. \ref{tab:SVMfeatures} reports a summary of the defined features.

\subsubsection*{Classification based on CNN}

The architecture of the chosen CNN is shown in Fig. \ref{fig:WinCoreNet_architecture}, and it has again been derived following the standard suggestions for convolutional networks \cite{DeepLearningBook2016}. It is composed of four 1D convolutional layers, each one doubling the number of channels of its input, followed by max-pooling, and in the last layer an adaptive average pooling to bring down the dimension of the time trace to one followed by a softmax classification layer. 
In the training phase, a dropout layer was also inserted before the average pooling, with a drop probability of 0.5. \\
The optimizer used for training is again Adam, with learning rate 0.001 and weight decay $10^{-5}$ and using as loss function the binary cross-entropy. The training phase was stopped once the training loss had been below a threshold (0.18) for at least 5 iterations.

\subsection{Selection based on identification results}

\begin{figure}[tb]
	\centering
		\subfloat{\includegraphics[width=1\columnwidth]{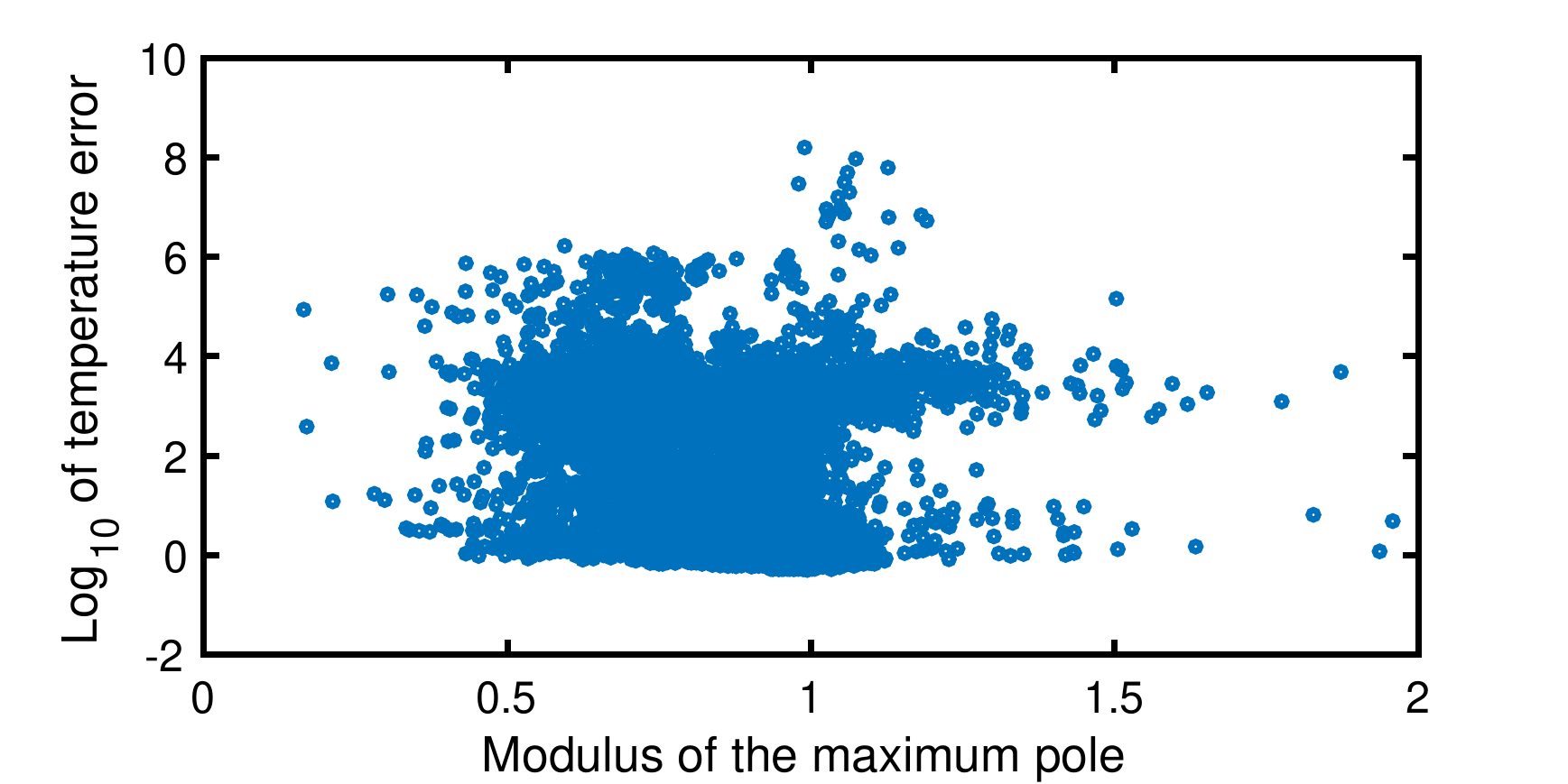}} \\
		\subfloat{\includegraphics[width=1\columnwidth]{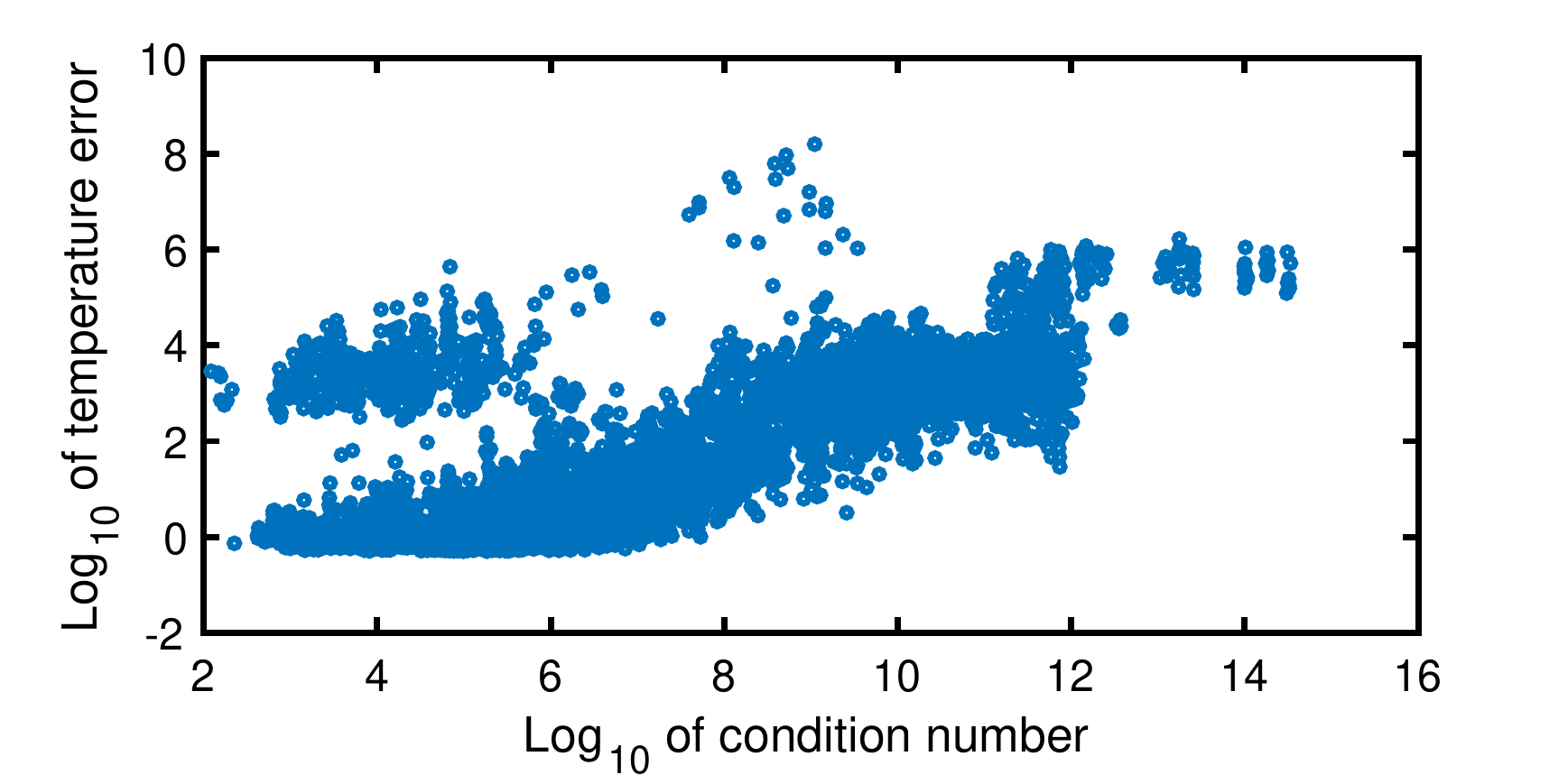}}
		\vspace*{-0.2cm}
			\caption{Correlation between the condition number or the maximum pole modulus and the temperature error for all cores and windows. For the magnitude of the maximum poles, we have calculated an average of $0.95\pm0.03$ for the good windows and of $0.85\pm0.17$ for the bad ones.}
	\label{fig:ThermmodIdentMetrics}
\end{figure}

In order to understand if the choice of a machine learning algorithm for the selection based on identification results is appropriate, we have started considering only simple metrics, i.e., the modulus of the maximum pole and the condition number of the matrix $\hat R$ for each core and window, where (see Sec. \ref{sec:therm_mod}):
\begin{equation*}
    \hat R = \begin{bmatrix} \hat\Sigma^T &
                             \hat\Sigma_q^T \end{bmatrix}^T. 
\end{equation*}
In fact, on the one hand, unstable models (where the poles magnitude is greater than 1) do not comply with the physics of the system and, on the other hand, we know from theoretical considerations that very high condition numbers are undesirable \cite{Ljung,SodSto}. As Fig. \ref{fig:ThermmodIdentMetrics} shows, these metrics can give very good indications to classify between good and bad windows, however there is still a large number of outliers. We have chosen therefore to resort to a machine learning algorithm, considering all the available metrics that we could derive from the identification algorithm.

\begin{table}[tb]
\caption{Chosen features for the SVM and NN algorithms using the results from the model identification.}
\vspace*{-0.2cm}
\label{tab:SVMIdentfeatures}
\begin{tabularx}{\columnwidth}{lX}
\hline
Feature definition        & Description \\ \hline
$\text{Re}\, p_i$ & Real parts of the poles of the $A$ matrix \\
$\text{Im}\, p_i$ & Imaginary parts of the poles of the $A$ matrix \\
$\sigma_w$ & Standard deviation of the equation error \\
$\sigma_v$ & Standard deviation of the measurement noise \\
$\text{corr}\left(res\right)$ & First four autocorrelations of the identification residuals \\
$\log_{10}\left(\text{cond}\left[\hat R\right]\right)$ & Condition number of the matrix $\hat R$\\
$\log_{10}\left(\min \text{svd}\left[\hat R\right]\right)$ & Minimum singular value of the matrix $\hat R$ \\ \\ \hline
\end{tabularx}
\end{table}

As already outlined, we have used again both a SVM algorithm with radial basis functions and a fully connected neural network. The architecture of the network is shown in Fig. \ref{fig:MLCoreIdentNet_architecture} and it has been derived very similarly to the one in Fig. \ref{fig:MLCoreNet_architecture}. It consists of four fully connected layers each followed by a rectified nonlinearity (ReLu) and finally a softmax classification layer. The optimizer used for training is once again Adam, with learning rate 0.001 and weight decay $10^{-4}$ and using as loss function the binary cross-entropy. The training phase was stopped once the training loss had been below a threshold (0.15) for at least 5 iterations. \\
Both algorithms use the same input, which consists in a set of features derived from the results of the identification algorithm. These features are reported in Tab. \ref{tab:SVMIdentfeatures}.

\subsection{Classification results}
\label{sec:class_results}

%

\begin{figure}[tb]
	\centering
		\subfloat[Time traces]{\includegraphics[width=1\columnwidth]{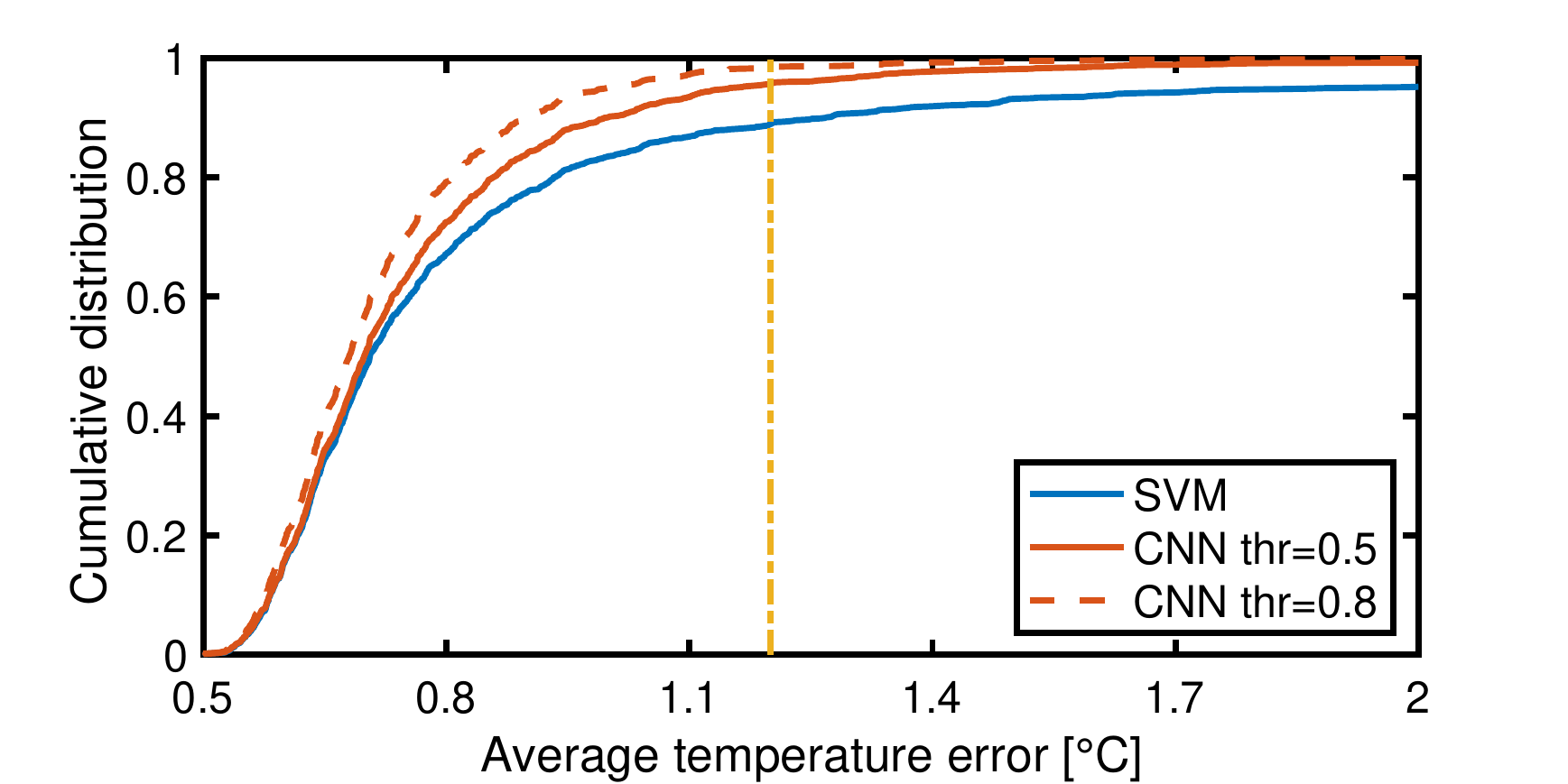} \label{fig:Tecdf_time}} \\
		\subfloat[Identification features]{\includegraphics[width=1\columnwidth]{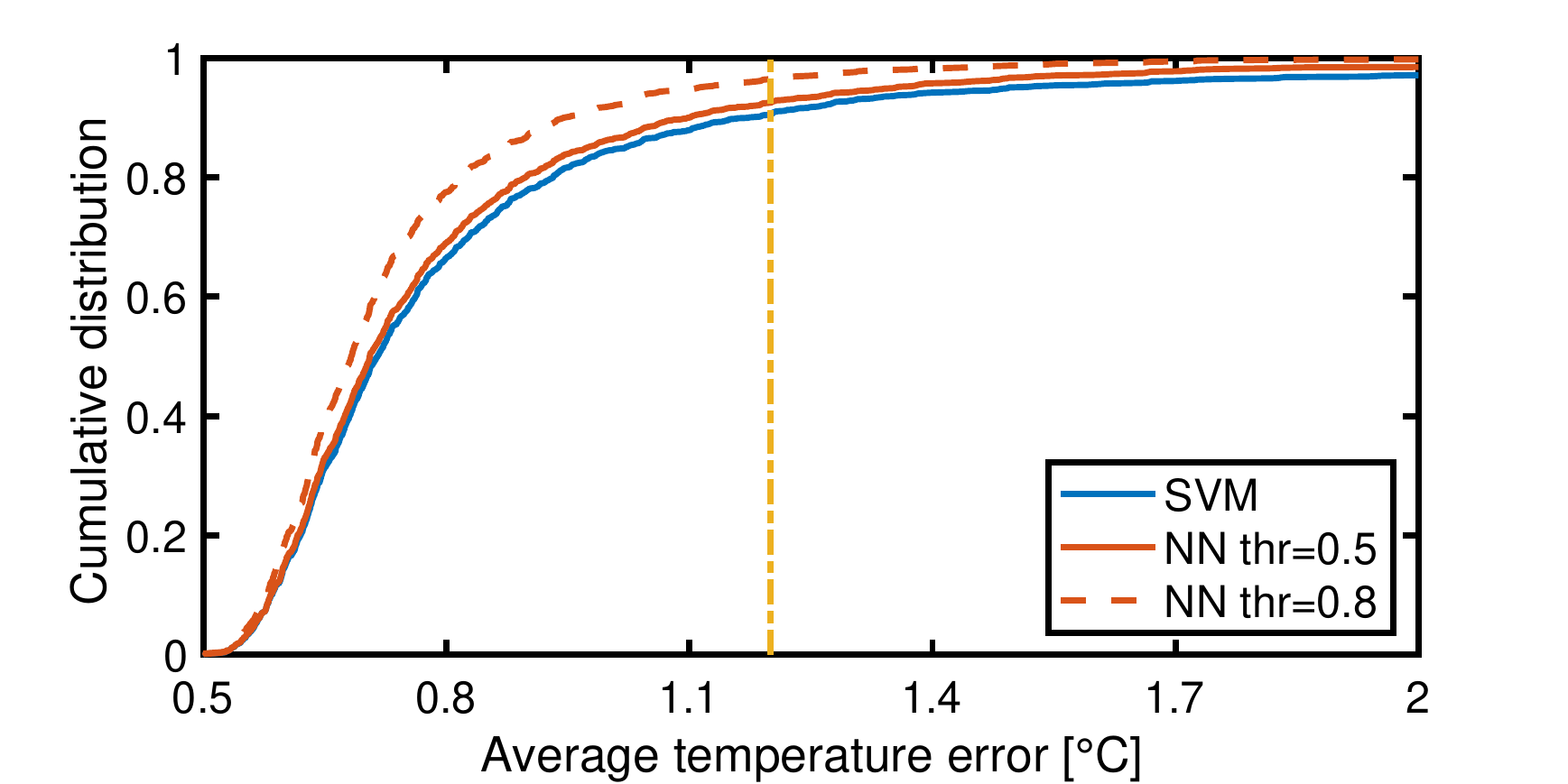} \label{fig:Tecdf_ident}} 
		\caption{Cumulative distributions (ECDFs) of the average temperature error for the windows chosen by the classification algorithm using either directly the time traces or the features coming from the identification, and varying the decision threshold for the neural networks. The chosen windows differ between the algorithms, so the ECDFs do not have the same number of points. The vertical dashed line is the 1.2$^\circ$C threshold.}
	\label{fig:Tecdf}
\end{figure}

\begin{table}[tb]
\centering
\caption{Classification accuracy for both algorithms, varying the decision threshold for the CNN. Note that all percentages are normalised by the number of classified windows, except for the last line (unclassified windows), which is normalised by the total number of windows in the test set.}
\vspace*{-0.2cm}
\label{tab:WinCore_classerr}
\begin{tabularx}{\columnwidth}{lllX}
\hline
\textbf{Metric definition}  & \textbf{Algorithm}  & \textbf{Threshold}  & \textbf{Value} \\ \hline
\multirow{8}{*}{\shortstack[l]{Percentage of misclassified \\good windows}} & SVM trace & - & 7.6\% \\
 & \multirow{2}{*}{NN trace} & 0.5 & 10\% \\
 &  & 0.8 & 3\% \\
 & \multirow{2}{*}{CNN trace} & 0.5 & 4.6\% \\
 &  & 0.8 & 2\% \\
 & SVM identification & - & 7.8\% \\
 & \multirow{2}{*}{NN identification} & 0.5 & 5\% \\
 &  & 0.8 & 1.8\% \\\hline
\multirow{8}{*}{\shortstack[l]{Percentage of misclassified \\bad windows}} & SVM trace & - & 6.8\% \\
 & \multirow{2}{*}{NN trace} & 0.5 & 6.8\% \\
 &  & 0.8 & 2.9\% \\
 & \multirow{2}{*}{CNN trace} & 0.5 & 3.5\% \\
 &  & 0.8 & 1.3\% \\
 & SVM identification & - & 3.2\% \\
 & \multirow{2}{*}{NN identification} & 0.5 & 4.2\% \\
 &  & 0.8 & 1.3\% \\\hline
\multirow{8}{*}{\shortstack[l]{Percentage of correctly \\classified windows}} & SVM trace & - & 86\% \\
 & \multirow{2}{*}{NN trace} & 0.5 & 83\% \\
 &  & 0.8 & 94\% \\
 & \multirow{2}{*}{CNN trace} & 0.5 & 91\% \\
 &  & 0.8 & 96\% \\
 & SVM identification & - & 89\% \\
 & \multirow{2}{*}{NN identification} & 0.5 & 90\% \\
 &  & 0.8 & 96\% \\\hline
\multirow{8}{*}{\shortstack[l]{Percentage of unclassified \\windows (relative to the \\size of the test set)}} & SVM trace & - & 0\% \\
 & \multirow{2}{*}{NN trace} & 0.5 & 0\% \\
 &  & 0.8 & 42\% \\
 & \multirow{2}{*}{CNN trace} & 0.5 & 0\% \\
 &  & 0.8 & 16\% \\
 & SVM identification & - & 0\% \\
 & \multirow{2}{*}{NN identification} & 0.5 & 0\% \\
 &  & 0.8 & 18\% \\ \hline
\end{tabularx}
\vspace*{-0.2cm}
\end{table}

Initially we concentrate on the study of the misclassified good windows, i.e., the cases where the algorithm classifies the window as being good while in reality it is not. Fig. \ref{fig:Tecdf_time} reports, for the SVM and CNN algorithms on the time traces and varying the decision threshold for the CNN, the cumulative distributions of the average temperature error on the entire data when the model has been identified in a window that is chosen by the algorithm as a good window. We immediately note that the CNN performs much better than the SVM. Moreover, if we raise the decision threshold to 0.8, the number of outliers is almost zero. \\
A similar situation can be observed in Fig. \ref{fig:Tecdf_ident}, which reports the same results for the SVM and NN algorithms on the identification results. Also in this case the SVM provides the lowest accuracy, while the NN with decision threshold 0.8 gives excellent performance. Moreover, comparing Fig. \ref{fig:Tecdf_time} and Fig. \ref{fig:Tecdf_ident}, we note that the CNN on time traces and the NN on identification results perform very similarly.

Finally, Tab. \ref{tab:WinCore_classerr} reports the percentages of correctly and incorrectly classified windows for all cases. It also shows the decrease in yield when the decision threshold is increased from 0.5 to 0.8. From the results reported in the table, we again conclude that the best performance is achieved with the CNN on time traces and the NN on identification results. Increasing the decision threshold to 0.8 allows much better performance, with the drawback of a relatively small decrease in yield. \\
Note that, even though the NN on identification results perform very similarly to the CNN on time traces, the former needs also to run the identification procedure before passing the results through the neural network, while the latter works directly on the raw signals.

\section{Conclusions}

In this work we have demonstrated the identification and application of a thermal model, suitable for power and thermal control and anomaly detection, to the nodes of an HPC cluster in production. We have shown that the performance of the model is excellent, with an average error on the temperature predicted using a Kalman filter very close to the quantization step. This result is even more relevant since the thermal models have been identified using real workloads on the nodes, and not ad-hoc excitations which would have required to put the production machine off-line. 

In order to achieve the best model performance, it is crucial to accurately choose the data to use for the model identification. To this purpose, we have analyzed and compared a variety of approaches based on machine learning and deep learning techniques that can reliably choose the appropriate windows of data given real workloads. Our finding is that this choice of window is indeed a non-trivial problem, which only sophisticated deep learning algorithms can accurately address. 

Our work paves the way to the application of thermal models on an HPC cluster in a scalable and efficient way, without requiring neither large computational overheads nor down-times of the machines for model calibration.

\section{Acknowledgements}

This work was supported by the EU FETHPC project ANTAREX (g.a. 671623). The authors would like also to thank Andrea Borghesi from the University of Bologna for providing the numbers on the usage of the cluster and Francesco Beneventi from the University of Bologna for his support on the framework and on the data collection.

\bibliographystyle{IEEEtran}
\bibliography{bibliography}

\end{document}